\documentclass{article}

\usepackage{nips11submit_e,times}

\usepackage{amssymb}
\usepackage{amsmath}
\usepackage{amsrefs}
\usepackage{xifthen}
\usepackage{tikz}
\usepackage{pgfplots}
\usepackage{todonotes}
\usepackage{soul}
\usepackage{subfigure}
\usepackage{mathdots}

\usepackage{graphics}

\def\sesc#1{%
\if#1.-%   . -> -
\else%
\if#1=-%   = -> -
\else%
\if#1,-%   , -> -
\else%
\if#1/-%   / -> -
\else#1%
\fi%
\fi%
\fi%
\fi%
}

\def\escnull{}

\def\esc#1#2\escend{%
\ifx#1\escnull%
\else%
\sesc#1\esc#2\escend%
\fi%
}

\def\escape#1{\esc#1\escnull\escend}

\newcommand{\includetikz}[2][]{%
  \def\graphicname{#2-\escape{#1}}%
  \includegraphics{\graphicname}%
}%

\newcommand{\mbb}[1]{{%
\ifthenelse{\isin{#1}{abcdefghijklmnopqrstuvxyz}}%
  {{\mathbf #1}}%
  {\boldsymbol #1}%
}}
\newcommand{\mup}[1]{{%
\ifthenelse{\isin{#1}{abcdefghijklmnopqrstuvxyzABCDEFGHIJKLMNOPQRSTUVXYZ}}%
  {\text{#1}}%
  {#1}%
}}

\newcommand{\set}[1]{\mbb{#1}}
\newcommand{\seg}[1]{#1}
\newcommand{\ff}[1]{#1}

\newcommand{\cand}[1]{\set{#1}}
\newcommand{\asg}[1]{#1}

\newcommand{\trans}[0]{{\sf T}}

\def\vec#1{\mbb{#1}}

\title{Multi-Hypothesis CRF-Segmentation of Neural Tissue in anisotropic EM 
volumes}

\author{
Jan Funke\\
Institute of Neuroinformatics\\
University Z\"{u}rich/ETH Z\"{u}rich\\
\texttt{funke@ini.phys.ethz.ch} \\
\And
Bj\"{o}rn Andres\\
HCI, University Heidelberg\\
\texttt{bjoern.andres@hci.iwr.uni-heidelberg.de} \\
\AND
Fred A. Hamprecht\\
HCI, University Heidelberg\\
\texttt{fred.hamprecht@hci.iwr.uni-heidelberg.de} \\
\And
Albert Cardona\\
Institute of Neuroinformatics\\
University Z\"{u}rich/ETH Z\"{u}rich\\
\texttt{acardona@ini.phys.ethz.ch} \\
\And
Matthew Cook\\
Institute of Neuroinformatics\\
University Z\"{u}rich/ETH Z\"{u}rich\\
\texttt{cook@ini.phys.ethz.ch} \\
}

\nipsfinalcopy % Uncomment for camera-ready version

\begin{document}
\maketitle
\maketitle
\begin{abstract}
We present an approach for the joint segmentation and grouping of similar 
components in anisotropic 3D image data and use it to segment neural tissue in 
serial sections electron microscopy (EM) images.
We first construct a nested set of neuron segmentation hypotheses for each 
slice. A conditional random field (CRF) then allows us to evaluate both the 
compatibility of a specific segmentation \emph{and} a specific inter-slice 
assignment of neuron candidates with the underlying observations.  The model is 
solved optimally for an entire image stack simultaneously using integer linear 
programming (ILP), which yields the maximum \emph{a posteriori} solution in 
amortized linear time in the number of slices.
%
%In each slice different segmentation hypotheses are enumerated.  Based on these 
%hypotheses, a conditional random field (CRF) rating all possible assignments of 
%similar components across slices is constructed.  Maximum \emph{a posteriori} 
%(MAP) inference on this CRF yields both the optimal segmentation of each slice 
%\emph{and} the assignment of components between slices. We find the globally 
%consistent and optimal solution in amortized linear time in the number of 
%slices. This is achieved by formulating the MAP inference task as an integer 
%linear program (ILP).
%
We evaluate the performance of our approach on an annotated sample of the 
\emph{Drosophila} larva neuropil and show that the consideration of different 
segmentation hypotheses in each slice leads to a significant improvement in the 
segmentation and assignment accuracy.
%
%\keywords{?,?,?}
%
\end{abstract}

\section{Introduction}
Electron microscopy (EM) remains the only imaging technique with sufficient 
resolution for the elucidation of synaptic contacts between 
neurons~\cite{Briggman2006}. The acquisition of large volumes of brain circuitry 
is now possible with recent advances in automated imaging for EM. The next 
bottleneck in the reconstruction of neural circuits is the accurate 
reconstruction of 3D neural arbors from stacks of EM images \cite{Jain2010}.  
Despite several efforts at automating the analysis of these highly stereotyped 
images, further improvements in overall accuracy are needed before automated 
methods can eliminate the tedious work currently needed to annotate large EM 
image datasets.

Current attempts at automatic labelling of anisotropic neural tissue in 
3D-imaged serial sections can broadly be divided in two categories. (i) Binary 
segmentation based approaches try to identify the outlines of neurons within 
single slices~\cite{Jurrus2010,Kaynig2010,Kaynig2010a}, and the results are used 
to establish geometrically consistent assignments of connected components that 
belong to one neuron. These approaches suffer from a high sensitivity to small 
errors:  a missing piece of membrane alters the topological properties of the 
result.  (ii) Over-segmentation based approaches merge small image regions 
within and between slices~\cite{Vazquez-Reina2010,Vitaladevuni2010}.  The 
resulting optimization problem is solved approximately.

Our approach belongs to the first category. The improvements over current work 
are: (i) Our framework allows for rivaling concurrent segmentation hypotheses.  
This reduces the likelihood of missing crucial parts of the segmentation. (ii) 
All possible continuations of neural processes across slices are considered 
jointly with the segmentation hypotheses.  Thus, higher-order relations between 
slices influence the segmentation. (iii) A globally consistent and optimal 
solution is found in amortized linear time in the number of slices. This is a 
prerequisite for the processing of large volumes.

\begin{figure}[t]
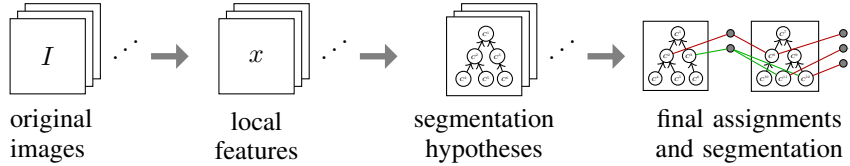

  \centering
  \includetikz{images/tikz/pipeline}
  \caption{The proposed processing pipeline.}
  \label{fig:pipeline}
\end{figure}

An overview of our approach can be seen in Fig.~\ref{fig:pipeline}. From the 
original images, per-pixel local features are extracted. Using several 
segmentations based on predictions of a pre-trained random forest classifier on 
the local features, \emph{segmentation hypotheses} are extracted for each slice.  
These hypotheses are connected components of pixels (see 
Sec.~\ref{sec:hypotheses}). Given the hypotheses, two tasks have to be 
performed. Firstly, a subset of non-overlapping components has to be found to 
make up the segmentation of each slice. Secondly, assignments have to be 
established between the selected components of two subsequent slices.  These 
assignments identify components that belong to the same neuron. We show how both 
tasks can be solved jointly by the introduction of binary \emph{assignment 
variables} (see Sec.~\ref{sec:assignment_model}). A consistent and optimal 
segmentation and assignment can be found by MAP-inference in a conditional 
random field (CRF).  Results presented in Sec.~\ref{sec:results} show that our 
approach leads to a significant improvement in the segmentation and assignment 
accuracy.

\section{Generating Segmentation Hypotheses}
\label{sec:hypotheses}

To extract segmentation hypotheses for each slice, we use a parameterized model 
to solve a binary image segmentation task. Given a field of per-pixel local 
features $\ff{x}$, the task is to maximize the conditional probability over a 
binary segmentation $\seg{y}$:
\begin{equation}
  p(\seg{y}|\ff{x})
  =
  \frac{1}{Z(\ff{x})}
  \exp\left[
    \lambda_D
    \sum_{i\in\Omega}
      D(\ff{x}_i,\seg{y}_i)
    -
    \lambda_S
    \sum_{i,j\in\mathcal{N}}
      S(\seg{y}_i,\seg{y}_j,\ff{x})
    -
    \lambda_N
    \sum_{i\in\Omega}
      \seg{y}_i
  \right]
  \label{eq:segmentation}
\end{equation}
Here, $\Omega \subset \mathbb{R}^2$ is the image domain and 
$D(\ff{x}_i,\seg{y}_i) = log\,p(\ff{x}_i|\seg{y}_i)$ is the log-likelihood of 
pixel $i\in\Omega$ belonging to fore- or background, as given by a pre-trained 
random forest classifier. The set $\mathcal{N} \subset \Omega\times\Omega$ 
contains all pairs of 8-connected neighboring pixels. $Z(\ff{x})$ is the 
partition function.  The second term in the exponent ensures smoothness of the 
segmentation, while favoring label changes in image regions of strong spatial 
gradients~\cite{Kohli2010}:
\begin{equation}
  S_{ij}(\seg{y}_i,\seg{y}_j,\ff{x})
  =
  \frac{
    \exp
    \left[
    -
    \frac{g^2(\ff{x}_i,\ff{x}_j)}
         {2\sigma^2}
    \right]
  }
  {
    \lVert i - j \rVert
  }
  (1-\delta_{\seg{y}_i = \seg{y}_j})
  \text{.}
\end{equation}
Here, $g(\ff{x}_i,\ff{x}_j)$ measures the difference in gray-levels\footnote{For 
simplicity, we assume that the gray level is part of the local feature vector 
$\ff{x}_i$.}, $\lVert .  \rVert$ denotes the length of a vector, and $\delta$ is 
the Kronecker-delta.  $\sigma$ is a parameter of the smoothness term.

\begin{figure}[t]
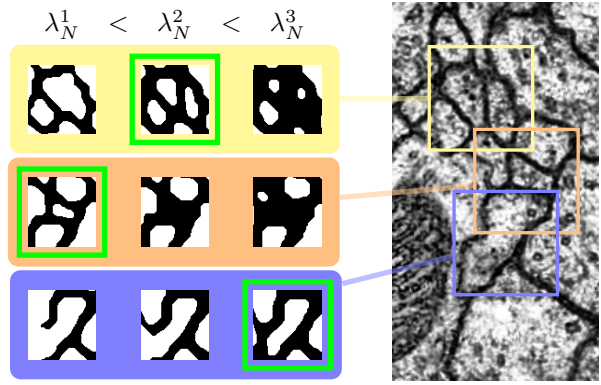

  \centering
  \includetikz{images/tikz/claim}
  \caption{Segmentation results of nearby image regions for different values of 
$\lambda_N$. This parameter determines how many of the pixels are getting 
assigned to neurons. The desired segmentation result for each case is 
highlighted in green.}
  \label{fig:claim}
\end{figure}

\begin{figure}
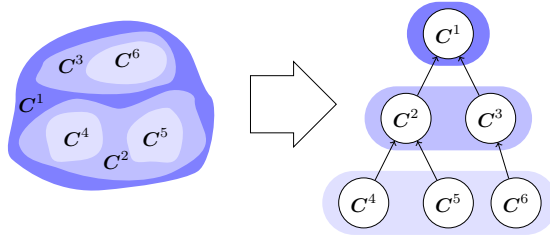

  \centering
  \includetikz[scale=0.75]{images/tikz/component_tree}
  \caption{Visualisation of the segmentation hypotheses extraction. For 
different values of the prior parameter $\lambda_N$ (shades of blue) connected 
components of the segmentation are found (left side). The subset relation of 
these connected components define the component tree (right side).}
  \label{fig:component_tree}
\end{figure}

The last term in the exponent of (\ref{eq:segmentation}) is a prior on the 
expected number of pixels assigned to the label `neuron'. The scalars 
$\lambda_D$, $\lambda_S$, and $\lambda_N$ are parameters of the probability 
distribution.  Since the exponent in (\ref{eq:segmentation}) is submodular, the 
inference task can be seen as a parametric max-flow problem which can be solved 
efficiently~\cite{Boykov2004}. However, finding an optimal set of parameters is 
a non-trivial task and one cannot expect a fixed set of parameters to perform 
well on all images~\cite{Kolmogorov2007}.  We go further and claim that a fixed 
set of parameters cannot even be expected to perform well on all areas of one 
image.  Therefore, we enumerate several local segmentation hypotheses by 
variation of $\lambda_N$ (Fig.~\ref{fig:claim}). This can be done efficiently by 
several warm-started graph-cuts~\cite{Kohli2010}, from which we obtain a series 
of segmentations.  Each segmentation consists of a set of connected components 
$\cand{C}^i \subset \Omega$ that are labelled as neurons. Each of these 
components is considered as one segmentation hypothesis. As $\lambda_N$ 
decreases, these components can grow and merge, thus establishing a tree-shaped 
subset hierarchy, the so-called \emph{component tree}~\cite{Jones1997} 
(Fig.~\ref{fig:component_tree}). All components which do not meet a criterion 
for stability over the range of values of $\lambda_N$ are 
removed~\cite{Matas2004}.  Let $\mathcal{C}^z$ denote the set of all remaining 
components of slice $z$. Any consistent subset $\mathcal{S}^z \subset 
\mathcal{C}^z$ of these hypotheses yields a valid segmentation of this slice. A 
subset is consistent if none of the containing components overlap, \emph{i.e.}, 
$\cand{C}^i\cap\cand{C}^j=\varnothing$ for all $\cand{C}^i,\cand{C}^j \in 
\mathcal{S}^z$ with $i\neq j$.  In the following, we show how we find the 
optimal consistent segmentation by considering the assignments of hypotheses 
between pairs of adjacent slices.
%
%The result of the variation of $\lambda_N$ is a sequence of segmentations.  
%Between the two extremal cases, in which all pixels are assigned to membranes 
%and all pixels are assigned to neurons, we observe the evolution of connected 
%components $\cand{C}^i \subset \Omega$ that are labelled as neurons: as 
%$\lambda_N$ decreases, more and more pixels are getting assigned to this label.  
%Connected components can emerge, grow, or merge with neighbors, thus 
%establishing a tree-shaped subset hierarchy, the so-called \emph{hypotheses
%tree}. Components that do not meet a criterion for stability over the range of 
%values of $\lambda_N$ are removed~\cite{Matas2004}. Let $\mathcal{C}$ denote the 
%set of all remaining hypotheses of all slices.
%

\section{Assignment Model}
\label{sec:assignment_model}

For each possible assignment of a segmentation hypothesis in one slice to a 
hypothesis in the previous or next slice, we introduce one binary assignment 
variable. This variable is set to $1$ if the involved hypotheses and their 
mutual assignment are accepted.

Let $m$ be the number of all possible assignments. A binary vector $\vec{a} \in 
\{0,1\}^m$ of assignment variables is created similarly to the method proposed 
in~\cite{Padfield2010}.  Each possible continuation of a hypothesis $\cand{C}^i$ 
in slice $z$ to a hypothesis $\cand{C}^j$ in slice $z+1$ is represented by a 
variable $\asg{a}^{i\rightarrow j}$. A split of $\cand{C}^i$ in slice $z$ to 
$\cand{C}^j$ and $\cand{C}^k$ in slice $z+1$ is represented as 
$\asg{a}^{i\rightarrow j,k}$.  Similarly, each possible merge is encoded as 
$\asg{a}^{i,j\rightarrow k}$.  Appearances and disappearances of neurons are 
encoded as assignments to a special end node $\cand{E}$, i.e., for each 
hypothesis $\cand{C}^i$ we introduce two variables $\asg{a}^{i\rightarrow 
\cand{E}}$ and $\asg{a}^{\cand{E}\rightarrow i}$.  For the possible assignments, 
only components within a threshold distance to each other are considered. Thus, 
the number of assignment variables is linear in the number of segmentation 
hypotheses.  See Fig.~\ref{fig:assignments} for examples of assignments of a 
single segmentation hypothesis. We find the optimal segmentation and assignment 
via MAP-inference on the following CRF:
\begin{figure}[t]
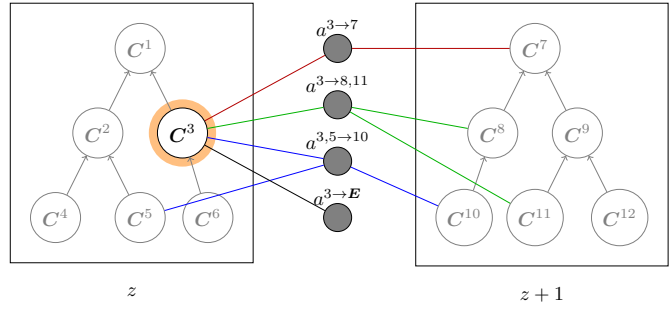

  \centering
  \includetikz[scale=0.75]{images/tikz/assignments}
  \caption{Examples of the four outgoing assignment categories for a single 
segmentation hypothesis (orange): A continuation (red) is modelled for each 
hypothesis in the next slice that is within a threshold distance. Possible 
splits (green) and merges (blue) are enumerated for neighboring hypotheses in 
the respective slices. Again, the possible sources or targets of splits and 
merges are segmentation hypotheses within a threshold distance.  The 
disappearance of a neuron is represented by a single assignment for each 
candidate (black).}
\label{fig:assignments}
\end{figure}
\begin{equation}
  p(\vec{a}|\mathcal{C})
  =
  \frac{1}{Z(\mathcal{C})}
  \exp
  \left[
    -
    \vec{c}^\trans\vec{a}
    -
    \sum_{\set{P}\in\mathcal{P}}
    H(\set{P})
    -
    \sum_{1\leq i \leq n}
    E(
      \vec{a}^{\rightarrow i},
      \vec{a}^{i\rightarrow}
    )
  \right]
  \text{.}
  \label{eq:crf}
\end{equation}
Here, $\mathcal{P}$ denotes the set of all complete paths of the hypotheses 
trees. The set of all \emph{incoming} and \emph{outgoing} assignment variables 
for a hypothesis $\cand{C}^i$ is denoted by $\vec{a}^{\rightarrow i}$ and 
$\vec{a}^{i\rightarrow}$, respectively. Incoming assignment variables of a 
hypothesis $\cand{C}^i$ are all components of $\vec{a}$ that have $\cand{C}^i$ 
on the right side of the superscript. Outgoing assignment variables are defined 
analogously.

The first term in the exponent accounts for unary potentials determining the 
costs for each assignment. For that, a vector $\vec{c} \in \mathbb{R}^m$ is 
constructed, which is congruent to $\vec{a}$. The costs for one-to-one 
assignments are modelled as:
\begin{equation}
  c^{i\rightarrow j}
  =
  \theta_L
    L(\cand{C}^{ij})
  +
  \theta_P
    \lVert
      \overline{\cand{C}^i} - \overline{\cand{C}^j}
    \rVert^2
  +
  \theta_S
    \lvert
      \cand{C}^i\ominus\cand{C}^j
    \rvert^2
  \text{.}
\end{equation}
Here, we write $\cand{C}^{ij}$ as a shorthand for $\cand{C}^i\cup\cand{C}^j$.  
The expression $\overline{\cand{C}^i}$ denotes the mean pixel position of the 
candidate $\cand{C}^i$; $\cand{C}^i\ominus\cand{C}^j$ is the mean-corrected 
symmetric set difference of candidates $\cand{C}^i$ and $\cand{C}^j$; $\lVert .  
\rVert$ denotes the Euclidean distance; and $\lvert . \rvert$ the cardinality of 
a set. The term $L(\cand{C})$ is proportional to the probability of assigning 
all pixels of $\cand{C}$ to a neuron, {\it i.e.},
\begin{equation}
  L(\cand{C})
  =
  \sum_{j \in \cand{C}}
  \left(
    D(\ff{x}_j,0)
    -
    D(\ff{x}_j,1)
  \right)
  \;
  +
  \sum_{j,k \in \mathcal{N};
        \atop
        j \in    \cand{C},\;
        k \notin \cand{C}}
    S_{j,k}(\seg{y}_j,\seg{y}_k,\ff{x})
  \text{.}
\end{equation}
In a similar way, we define the costs for splits:
\begin{align}
  c^{i\rightarrow j,k}
  &=
  \theta_L
    L(\cand{C}^{ijk})
  +
  \theta_{BP}
    \lVert
      \overline{\cand{C}^i} - \overline{\cand{C}^{jk}}
    \rVert^2
  +
  \theta_{BS}
    \lvert
      \cand{C}^i\ominus\cand{C}^{jk}
    \rvert^2
  \text{.}
\end{align}
The merge cases are defined analogously. Costs for the appearance or 
disappearance of a neuron depend on the data term and the size of the component:
\begin{align}
  c^{i\rightarrow \cand{E}}
  =
  c^{\cand{E}\rightarrow i}
  =
  \theta_L
    L(\cand{C}^i)
  +
  \theta_E
    \lvert
      \cand{C}^i
    \rvert^2
  \text{.}
\end{align}
The scalars $\theta_L$, $\theta_P$, $\theta_S$, $\theta_{BP}$, 
$\theta_{BS}$, and $\theta_E$ are parameters of the model.

\begin{figure}[t]
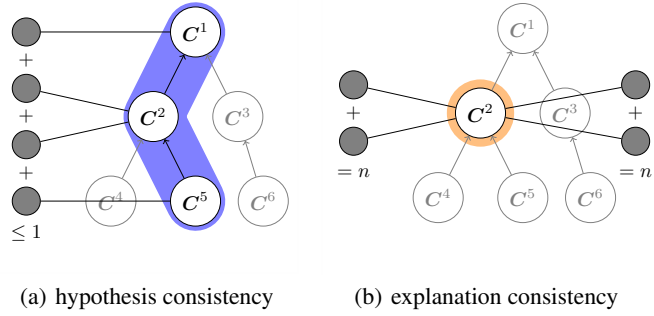

\centering
\subfigure[hypothesis consistency]{
  \includetikz[scale=0.75]{images/tikz/consistency_hypotheses}
  \label{fig:consistency:hypotheses}
}
\subfigure[explanation consistency]{
  \includetikz[scale=0.75]{images/tikz/consistency_explanation}
  \label{fig:consistency:explanation}
}
\caption{Visualization of the two types of consistency constraints. The 
hypothesis consistency \subref{fig:consistency:hypotheses} ensures that no pixel 
is assigned to more than one segmentation hypothesis: For each path of the 
component tree (blue), the sum of all \emph{incoming} assignment variables 
(gray) has to at most one.  The explanation consistency 
\subref{fig:consistency:explanation} ensures a continuous sequence of 
assignments: For each segmentation hypothesis (orange), the sum of all 
\emph{incoming} assignment variables (from the previous image) has to be equal 
to the sum of all \emph{outgoing} assignment variables (to the next image). The 
incoming or outgoing assignment variables for a component are all assignment 
variables that have the component as target or source, respectively.}
\end{figure}

The second and third terms in (\ref{eq:crf}) are higher-order potentials that 
ensure the consistency of the solution. $H(\set{P})$ ensures that at most one of 
all incoming assignments of competing segmentation hypotheses is selected.  
Competing hypotheses are components that share some pixels, i.e., all components 
along one path in the hypotheses tree (see 
Fig.~\ref{fig:consistency:hypotheses}).  This \emph{hypothesis constraint} 
ensures that each pixel is explained by at most one hypothesis and that at most 
one of the possible assignments is selected for each hypothesis.
\begin{equation}
  H(\set{P}) =
  \left\{
    \begin{array}{ll}
      0    & \text{if}\;
        \sum_{i\in\set{P}}
        %\lvert \vec{a}^{\rightarrow i} \rvert
        \sum_{a\in \vec{a}^{\rightarrow i}}
        a
        \leq 1
      \\
      \infty & \text{else}
    \end{array}
  \right.
\end{equation}
$E(\vec{a}^{\rightarrow i},\vec{a}^{i\rightarrow})$ ensures that if an incoming 
assignment was selected for a segmentation hypothesis, an outgoing assignment is 
selected as well. This \emph{explanation constraint} guarantees a consistent 
sequence of assignments (see Fig.~\ref{fig:consistency:explanation}).
\begin{equation}
  E(\vec{a}^{\rightarrow i},\vec{a}^{i\rightarrow})
  =
  \left\{
    \begin{array}{ll}
      0    & \text{if}\;
        %\lVert \vec{a}^{\rightarrow i} \rVert
        \sum_{a \in \vec{a}^{\rightarrow i}} a
        =
        %\lVert \vec{a}^{i\rightarrow} \rVert
        \sum_{a' \in \vec{a}^{i\rightarrow}} a'
      \\
      \infty & \text{else}
    \end{array}
  \right.
\end{equation}

Since the potentials $H$ and $E$ impose hard constraints and the remaining 
potentials are linear in $\vec{a}$, the MAP solution to (\ref{eq:crf}) can be 
found by solving the following linear program.
\begin{align}
  \min \vec{c}^\trans\vec{a}
  %\\
  \hspace{1.5cm}
  \text{s.t.}
  \hspace{0.5cm}
  \sum_{i\in\set{P}}
  \sum_{a\in \vec{a}^{\rightarrow i}}
  a
  &
  \leq 1
  &
  \forall \set{P} \in \mathcal{P}
  \\
  \sum_{a \in \vec{a}^{\rightarrow i}} a
  -
  \sum_{a' \in \vec{a}^{i\rightarrow}} a'
  &
  = 0
  &
  1 \leq i \leq n
\end{align}

Unfortunately, the constraint matrix of this linear program is not totally 
unimodular. Therefore, we have to enforce the integrality of the solution 
explicitly. The resulting optimization problem is an instance of an integer 
linear program (ILP), which we solve using the IBM CPLEX solver~\cite{cplex}.

\section{Results}
\label{sec:results}

We evaluated the performance of our approach on an annotated sample of 
\emph{Drosophila} larva neural tissue~\cite{Cardona2010}. This publicly 
available data set consists of 30 serial sections (50 nm), imaged with 
transmission electron microscopy at a resolution of 4x4x50 nm/pixel. The image 
volume contains a 2x2x1.5 micron cube of neuropil tissue. The dataset includes 
labels of cellular membranes, cytoplasms and mitochondria of all 170 neural 
processes contained in the dataset.

To measure the accuracy of our method, we use the edit distance between the 
result and the ground-truth, i.e., the number of splits and merges a human 
operator would need to perform to restore the 
ground-truth~\cite{Vitaladevuni2010}. In contrast to the measure proposed 
in~\cite{Jain2010}, we count every false merge, even if the same objects are 
involved. Each missed neuron segment in a slice is counted as one merge error 
and each falsely introduced neuron segment is counted as one split error.  
Between the slices, we count each missed assignment as a split error and each 
falsely introduced assignment as a merge error.

\begin{figure}[t]
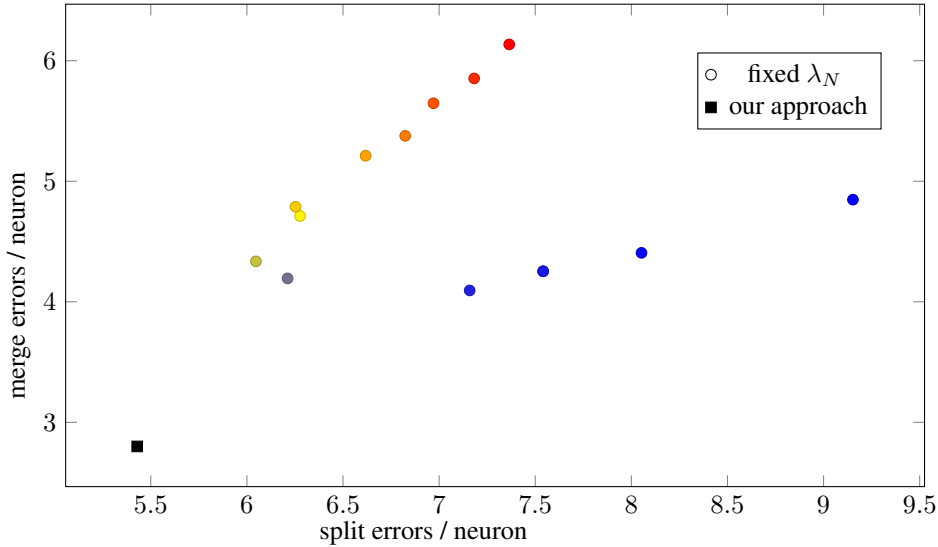

  \centering
  \def\nneurons{170}
  \def\nslices{1}
  \def\plotwidth{13cm}
  \def\plotheight{8cm}
  \includetikz{images/plots/pgf/single_thresholds}
  \caption{Comparison of the accuracy of our approach with different 
  segmentation hypotheses (black square) against a series of experiments without 
segmentation hypotheses (colored circles). The consideration of different 
segmentations is superior to all possible single segmentations. The single 
segmentation experiments cover the whole range from under-segmentation (blue) to 
oversegmentation (red).}
  \label{fig:single_thresholds}
\end{figure}

To evaluate the performance of our approach in comparison to approaches that do 
not allow local variations in the segmentation parameters (which is the case for 
all existing approaches that we know of), we carried out a series of experiments 
with fixed $\lambda_N$ (see Eq.~\ref{eq:segmentation}).  In particular, we took 
34 equidistant samples of $\lambda_N$ in an interval reaching from obvious over- 
to under-segmentation of the images. Note that the clamping of the parameter 
$\lambda_N$ corresponds to not considering conflicting hypotheses: all the 
component trees have a depth of one. All other parameters of the pipeline do not 
change between the experiments, and have been found via a grid-search in the 
parameter space.  We compare the results of this series with a single run of our 
proposed method that does allow for local variations of $\lambda_N$. In 
Fig.~\ref{fig:single_thresholds} we show the edit distance for each $\lambda_n$ 
in the series as well as for our approach normalized by the number of neurons in 
the dataset. Our data demonstrates that the single run with local variation of 
$\lambda_N$ is superior to all the cases in which $\lambda_N$ was fixed.

\begin{figure}[p]
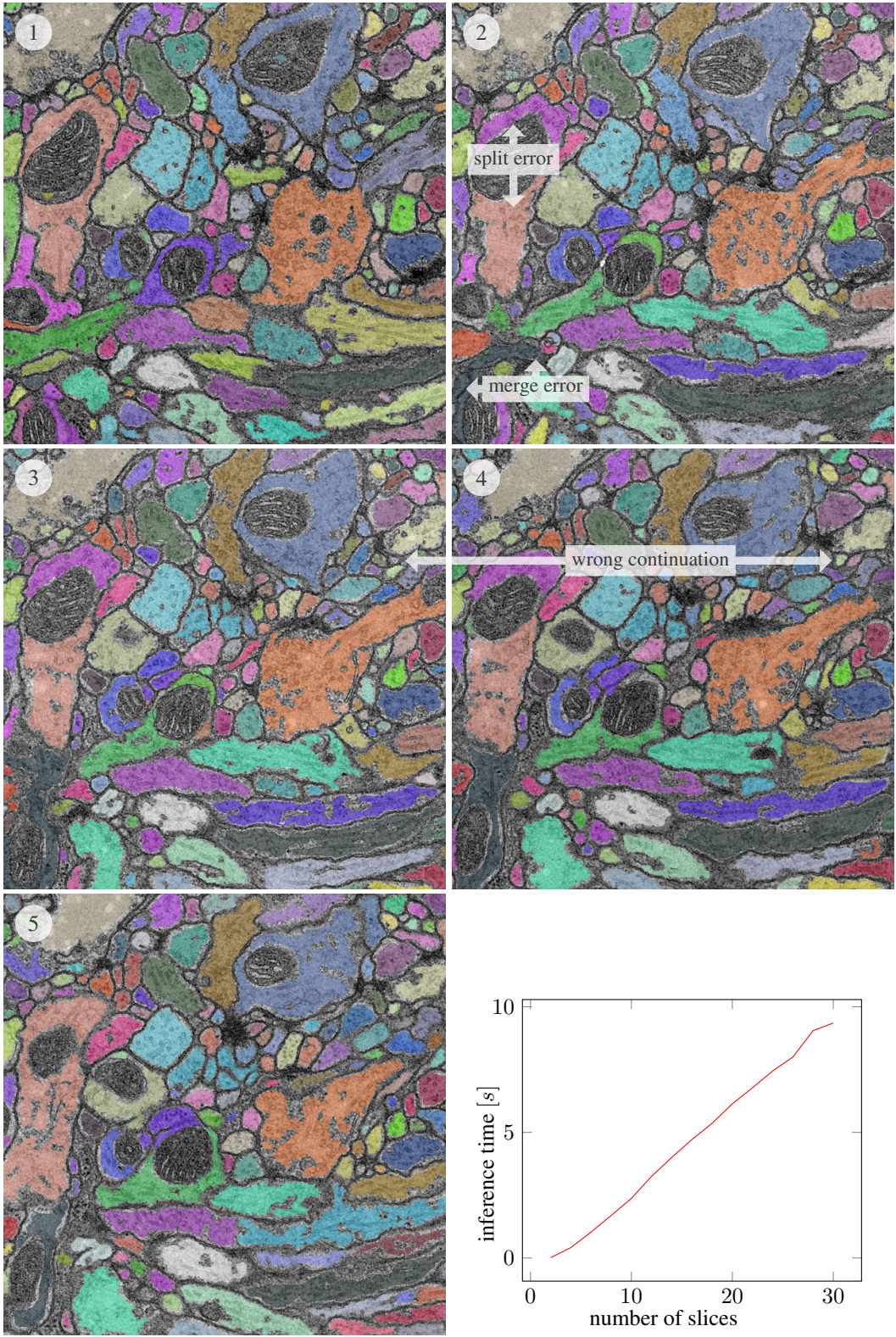

  \includetikz{images/tikz/best_result_runtime}
  \caption{Sample of the segmentation result of five subsequent slices (a) and 
  inference time (b). Examples of each error type are highlighted. Colors 
indicate the assignment of components between the slices.}
  \label{fig:result_sample}
\end{figure}

In Fig.~\ref{fig:result_sample} we show a representative segmentation example of 
five subsequent slices using our approach. In the same figure we give the 
inference time for the processing of the test dataset with different number of 
slices. The results indicate that the solution to the optimization task can be 
found in amortized linear time.

\section{Conclusion}
\label{sec:conclusion}

We presented a novel approach for the joint segmentation and assignment of 
similar regions in anisotropic 3D image data with an amortized linear running 
time in the number of slices. On an annotated sample of neural tissue we could 
show that the consideration of different segmentation hypotheses significantly 
improves the accuracy of the segmentation and grouping of neural processes. The 
probabilistic formulation of our approach is particularly useful in 
semi-interactive environments, where human-reconstructed neurons can help 
resolving ambiguity. Every manually traced neural process can be used to impose 
strong priors on the assignment model by decreasing the costs for the respective 
assignments.

Currently, the assignment costs are computed from simple statistics of the 
involved segmentation hypotheses.  Improvements might be obtained by using more 
sophisticated features~\cite{Kaynig2010}. This will be the focus of further 
research.

\section{Acknowledgements}

This work was funded by the Swiss National Science Foundation.
Parts of this work have been completed in collaboration with the Heidelberg 
Collaboratory for Image Processing.

\bibliography{references}

\begin{bibdiv}
\begin{biblist}

\bib{cplex}{misc}{
       title={{IBM} {ILOG} {CPLEX} {O}ptimizer, v12.2},
  note={\url{http://www-01.ibm.com/software/integration/optimization/cplex-opt%
imizer/}},
}

\bib{Boykov2004}{article}{
      author={Boykov, Yuri},
      author={Kolmogorov, Vladimir},
       title={{A}n experimental comparison of min-cut/max- flow algorithms for
  energy minimization in vision},
        date={2004sept.},
        ISSN={0162-8828},
     journal={{IEEE} {T}ransactions on {P}attern {A}nalysis and {M}achine
  {I}ntelligence},
      volume={26},
      number={9},
       pages={1124 \ndash 1137},
}

\bib{Briggman2006}{article}{
      author={Briggman, Kevin~L},
      author={Denk, Winfried},
       title={{T}owards neural circuit reconstruction with volume electron
  microscopy techniques},
        date={2006},
     journal={{C}urrent {O}pinion in {N}eurobiology},
      volume={16},
      number={5},
       pages={562 \ndash  570},
}

\bib{Cardona2010}{article}{
      author={Cardona, Albert},
      author={Saalfeld, Stephan},
      author={Preibisch, Stephan},
      author={Schmid, Benjamin},
      author={Cheng, Anchi},
      author={Pulokas, Jim},
      author={Tomancak, Pavel},
      author={Hartenstein, Volker},
       title={{An integrated micro- and macroarchitectural analysis of the
  \emph{Drosophila} brain by computer-assisted serial section electron
  microscopy }},
        date={2010},
     journal={{PLoS Biology}},
      volume={8},
      number={10},
       pages={e100050},
}

\bib{Jain2010}{article}{
      author={Jain, Viren},
      author={Seung, H.~Sebastian},
      author={Turaga, Srinivas~C.},
       title={{Machines that learn to segment images: a crucial technology for
  connectomics}},
        date={2010},
     journal={{C}urrent {O}pinion in {N}eurobiology},
}

\bib{Jones1997}{conference}{
      author={Jones, R.},
       title={{C}omponent {T}rees for {I}mage {F}iltering and {S}egmentation},
        date={1997},
   booktitle={{P}roceedings of the 1997 {IEEE} {W}orkshop on {N}onlinear
  {S}ignal and {I}mage {P}rocessing. {M}ackinac {I}sland},
}

\bib{Jurrus2010}{article}{
      author={Jurrus, Elizabeth},
      author={Paiva, Antonio~R.C.},
      author={Watanabe, Shigeki},
      author={Anderson, James~R.},
      author={Jones, Bryan~W.},
      author={Whitaker, Ross~T.},
      author={Jorgensen, Erik~M.},
      author={Marc, Robert~E.},
      author={Tasdizen, Tolga},
       title={{D}etection of neuron membranes in electron microscopy images
  using a serial neural network architecture},
        date={2010},
        ISSN={1361-8415},
     journal={{M}edical {I}mage {A}nalysis},
      volume={14},
      number={6},
       pages={770 \ndash  783},
}

\bib{Kaynig2010}{inproceedings}{
      author={Kaynig, Verena},
      author={Fuchs, Thomas},
      author={Buhmann, Joachim~M.},
       title={{G}eometrical {C}onsistent 3{D} {T}racing of {N}euronal
  {P}rocesses in ss{TEM} {D}ata},
        date={2010},
   booktitle={{P}roceedings of the {C}onference on {M}edial {I}mage {C}omputing
  and {C}omputer-{A}ssisted {I}ntervention},
      series={Lecture Notes in Computer Science},
      volume={6362},
   publisher={{S}pringer {B}erlin / {H}eidelberg},
       pages={209\ndash 216},
}

\bib{Kaynig2010a}{inproceedings}{
      author={Kaynig, Verena},
      author={Fuchs, Thomas},
      author={Buhmann, Joachim~M.},
       title={{N}euron {G}eometry {E}xtraction by {P}erceptual {G}rouping in
  ss{TEM} {I}mages},
        date={2010},
   booktitle={{P}roceedings of the {IEEE} {C}onference on {C}omputer {V}ision
  and {P}attern {R}ecognition},
     address={Los Alamitos, CA, USA},
       pages={2902\ndash 2909},
}

\bib{Kohli2010}{incollection}{
      author={Kohli, Pushmeet},
      author={Torr, Philip},
       title={{D}ynamic {G}raph {C}uts and {T}heir {A}pplications in {C}omputer
  {V}ision},
        date={2010},
   booktitle={{C}omputer {V}ision},
      series={Studies in Computational Intelligence},
      volume={285},
   publisher={{S}pringer {B}erlin / {H}eidelberg},
       pages={51\ndash 108},
}

\bib{Kolmogorov2007}{inproceedings}{
      author={Kolmogorov, Vladimir},
      author={Boykov, Yuri},
      author={Rother, Carsten},
       title={{A}pplications of {P}arametric {M}axflow in {C}omputer {V}ision},
        date={2007},
   booktitle={{P}roceedings of the {IEEE} {I}nternational {C}onference on
  {C}omputer {V}ision ({ICCV})},
}

\bib{Matas2004}{article}{
      author={Matas, Jiri},
      author={Chum, Ondrej},
      author={Urban, Martin},
      author={Pajdla, Tom\'{a}s},
       title={{R}obust {W}ide-{B}aseline {S}tereo from {M}aximally {S}table
  {E}xtremal {R}egions},
        date={2004},
        ISSN={0262-8856},
     journal={{I}mage and {V}ision {C}omputing},
      volume={22},
      number={10},
       pages={761 \ndash  767},
}

\bib{Padfield2010}{article}{
      author={Padfield, Dirk},
      author={Rittscher, Jens},
      author={Roysam, Badrinath},
       title={{C}oupled minimum-cost flow cell tracking for high-throughput
  quantitative analysis},
        date={2010},
        ISSN={1361-8415},
     journal={{M}edical {I}mage {A}nalysis},
      volume={In Press, Corrected Proof},
  url={http://www.sciencedirect.com/science/article/B6W6Y-50S8PV2-1/2/87a91a1d%
81d6904af9ca4059c373e0a7},
}

\bib{Vazquez-Reina2010}{inproceedings}{
      author={Vazquez-Reina, Amelio},
      author={Avidan, Shai},
      author={Pfister, Hanspeter},
      author={Miller, Eric},
       title={{M}ultiple {H}ypothesis {V}ideo {S}egmentation from {S}uperpixel
  {F}lows},
        date={2010},
   booktitle={{P}roceedings of the {E}uropean {C}onference on {C}omputer
  {V}ision ({ECCV})},
}

\bib{Vitaladevuni2010}{inproceedings}{
      author={Vitaladevuni, Shiv Naga~Prasad},
      author={Basri, Ronen},
       title={{C}o-{C}lustering of {I}mage {S}egments {U}sing {C}onvex
  {O}ptimization {A}pplied to {EM} {N}euronal {R}econstruction},
        date={2010},
   booktitle={{P}roceedings of the {IEEE} {C}onference on {C}omputer {V}ision
  and {P}attern {R}ecognition},
       pages={2203 \ndash 2210},
}

\end{biblist}
\end{bibdiv}
\end{document}